# A NOVEL IMAGE SEGMENTATION ENHANCEMENT TECHNIQUE BASED ON ACTIVE CONTOUR AND TOPOLOGICAL ALIGNMENTS

Ashraf A. Aly[1], Safaai Bin Deris[2] and Nazar Zaki[3]

[1, 2]Faculty of Computer Science, Universiti Teknologi, Malaysia

Ashraf.ahmed@uaeu.ac.ae, safaai@utm.my

[3]College of Information Technology, UAE University, UAE

nzaki@uaeu.ac.ae

## ABSTRACT

*Topological alignments and snakes are used in image processing, particularly in locating object boundaries. Both of them have their own advantages and limitations. To improve the overall image boundary detection system, we focused on developing a novel algorithm for image processing. The algorithm we propose to develop will based on the active contour method in conjunction with topological alignments method to enhance the image detection approach. The algorithm presents novel technique to incorporate the advantages of both Topological Alignments and snakes. Where the initial segmentation by Topological Alignments is firstly transformed into the input of the snake model and begins its evolvement to the interested object boundary. The results show that the algorithm can deal with low contrast images and shape cells, demonstrate the segmentation accuracy under weak image boundaries, which responsible for lacking accuracy in image detecting techniques. We have achieved better segmentation and boundary detecting for the image, also the ability of the system to improve the low contrast and deal with over and under segmentation.*

## 1. INTRODUCTION

Image segmentation is an important step in many techniques of multi-dimensional signal processing and its applications. Texture analysis is essential in many tasks such as shape determination, scene classification, and image processing. The segmentation process is known as determining the best positions of the image points according to the shape information. Algorithms based on classifiers have been applied to segment organs in medical images like cardiac and brain images. Image segmentation is a computer vision research area. It is the process of partitioning the digital image into meaningful regions with respect to a particular application. The goal of image segmentation is to cluster pixels into salient image regions. It is an initial and vital step in pattern recognition-a series of processes aimed at overall image understanding. Properties like gray level, colour, texture, shape help to identify regions and similarity of such properties, is used to build groups of regions having a particular meaning. The segmentation is based on measurements taken from the image and might be grey level, colour, texture, depth or motion. Applications of image segmentation include, identifying objects in a scene for object-based measurements such as size and shape, identifying objects in a moving scene for object-based video compression, and identifying objects which are at different distances from a sensor using depth measurements from a laser range finder enabling path planning for a mobile robots.

Segmentation could be used for object recognition, occlusion boundary estimation within motion or stereo systems, image compression, image editing, or image database look-up.
In mathematical sense the segmentation of the image *I*, which is a set of pixels, is the partition of *I* into *n* disjoint sets *R1,R2, . . . , Rn*, called segments or regions such that their union of all regions equals *I*,
I =$R_1$ U $R_2$ U….. U $R_n$ .

## 2. RELATED WORK

Accurate segmentation and boundary detection are important in microscopic imaging studies. For instance, Image analysis of leukocytes cells is essential part for curative and preventative treatments to many diseases and also important to understand and successfully treat inflammatory diseases as in Ray *et al.* [6]. Sensitive tracking for moving cells is important to do mathematical modelling to cell locomotion. Zimmer *et al.* [7] modified the active contour model to detect the mobility of the moving cells and also handle the cell division by providing an initial segmentation for the first frame. Mukherjee et al. [5] developed an algorithm by using threshold decomposition computed via image level sets to handle tracking problem and segmentation simultaneously. Li et al. [4] developed an algorithm with two levels, a motion filter and a level set tracker to handle the cell detection and the cells that move in and out of the image. Coskun et al. [2] used imaging data to solve the inverse modelling problem to determine the mobility analysis of the cells. Recently there have been a number of researchers attempt to create automated algorithms to detect and track the cells from microscopic images as in Mélange, et al. [18]; Mignotte et al. [19]; Krinidis et al. [20].

In this paper we managed to introduce a novel technique for image segmentation and cell boundary detection.

## 3. ALGORITHM

Topological alignments and snakes are used in image processing, particularly in locating object boundaries. Both of them have their own advantages and limitations. Active contour, can locate the object boundaries dynamically and automatically from an initial contour. The main advantage of active contour models is the ability of the method to give a piece wise linear description of the shape of the object at the time of convergence, without extra processing. .But the Active Contour models are strongly depends on finding strong image gradients to drive the contour. This significantly limits their utility, because frames with weak image boundaries and low contrast cause under and over segmentation which responsible for lacking accuracy. To solve this problem with the Active Contour model, we used Topological Alignments method to improve the performance of segmentation of cell tracking and increase the accuracy of cell tracking results. The method links segments between each frame and the next frame, reducing the number of false detections and false trajectories.

In this section, we propose a novel algorithm based on Topological Alignments and snake. We present novel image segmentation and tracking system technique based on the Active Contour method to be used in conjunction with Topological Alignments method to incorporate the advantages of both Topological Alignments and snakes for more accurate tracking approach to introduce a tracking system to detect and analyze the mobility of the living cells. It proceeds in two steps: in the first step, an initial segmentation by Topological Alignments method occurs to improve the performance of segmentation of cell tracking and improve cell tracking results. In the second step, transfer the output into the input of the snake model and begins its evolvement to the cells boundary and analyze the cells mobility.

We have tested the algorithm by using a gray scale images from Rodrigo et al. 2007 (Originally from CellAtlas.com1 reference library, a public cell image database) with images of human blood cells. We carried out tests on 70 images. The image processing algorithms are based on Open CV library (Bradskiet et al. [1]). We implemented the algorithm in C++ under Windows XP operating system.

## 3.1. Active Contour Models

The objective is to segment an image by deforming an initial contour towards the boundary of the object of interest. This is done by deforming an initial contour in such a way that it minimizes an energy functional defined on contours as in Kass et al. [3]; Ray et al. [6]; Zimmer et al. [7] and Sacan et al. [16]. The energy functional consists of two components: the first component is the potential energy, and it is small when the contour is aligned to the image edge, and the second component is the internal deformation energy, and it is small when the contour is smooth. The termination functional can be implemented with a gradient direction calculus in a slightly smoothed version of the image. Active contours models.

An Active contour can be parametrically represented by $v(s) = (x(s), y(s))$ and its energy functional can be written as:

$$E = \int_0^1 E_{int}(v(s))ds + \int_0^1 E_{image}(v(s))ds + \int_0^1 E_{ext}(v(s))ds \qquad (1)$$

Where $E_{int}$ represents the internal energy of the spline due to bending, $E_{image}$ gives rise to the image forces and $E_{ext}$ gives rise to the external constraint forces. The spline energy is controlled by $a(s)$ and $B(s)$. Therefore, the internal spline energy can be written as:

$$E_{int} = \frac{(a(s)|v_S(s)|^2 + B(s)|v_{SS}(s)|^2)}{2} \qquad (2)$$

The generic total image energy can be expressed as a weighted combination of the three energy functions.

$$E_{image} = w_{line}E_{line} + w_{edge}E_{edge} + w_{term}E_{term} \qquad (3)$$

The active contour models can be categorized into two models: region based models and edge-based models. The choice between these two types of models in applications depends on different characteristics of images. The main advantage of active contour models is the ability of the method to give a piece wise linear description of the shape of the object at the time of convergence, without extra processing. But the active contour models are strongly depends on finding strong image gradients to drive the contour. This significantly limits the use of the active contours.

## 3.2. Topological Alignments Method

The method links segmentation of two consecutive frames in the video sequence. Miura et al. [9]; Danuser et al.[17]; Zimmer et al.[8]; and Ersoy et al.[12]. From the output of the segmentation procedure, the method find the maximum weighted solutions between two pairs of frames. Then match the segments.

The method identify the segmentation of two images as $m$ and $n$ with an index set $P = \{1,..., m\}$, and an index set $Q = \{1,..., n\}$.

The method assume that cells move moderately between two consecutive frames, we assign the relative overlap of *p* and *q* as their weight, formally defined as

$$w(p, q) := | A(p) \cup A(q) | / | A(p) \cup A(q) | \qquad (4)$$

The sets of segments that achieve a relative overlap close to one it is more likely to be considered as one cell, but a relative overlap close to zero indicates segment sets that do not constitute one cell. According to these weights, we can determine the topological alignment notion. We denote $PL(M)$ for the set of all $L$-partitioning`s of a finite set $M$, and $S$ as a family of sets and to identify the $L$ subsets as $S = (S1,..., SL)$.

This allows us to state the alignment as finding those partitioning $S$ and $T$ that realize the maximum in the target function. The topological alignments method improves the performance of segmentation of cell tracking by explicitly taking into account the inherent problems of under (one segment fully covers two cells) segmentation, while still allowing the detection of cell division. The method links segments between each frame and the next frame, reducing the number of false detections and false trajectories.

## 4. RESULTS AND DISCUSSION

In order to obtain validation of our approach, we have tested the algorithms by using a gray scale images from Rodrigo et al. 2007 (Originally from CellAtlas.com1 reference library, a public cell image database) with images of human blood cells. We carried out tests on 70 images. The image processing algorithms are based on Open CV library (Bradskiet et al. [1]). Figure 1 illustrates some results where the processing of the initial image with another important contribution for using  the Novel techniques which defined better contrast and regularize the image contours, leading to a better segmentation without leaking,  and accurate segmentation for cell images. The results show the advantages of using our novel techniques to enhance the image and the effect on detecting the cell contours with better and accurate segmentation.

The results indicate improvement in segmentation performance by using the topological Alignments, which leads to improve cell detection results. Our results indicate better segmentation and more accurate detecting and analyzing leukocytes cells, also the ability of the system to improve the low contrast, under and over segmentation as shown in Figure 1, Figure 2, Figure 3, and Figure 4.

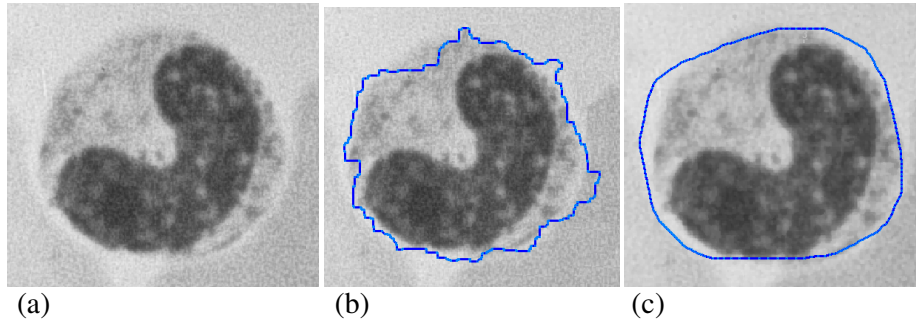

(a)                            (b)                            (c)

Figure 1. Example image from grayscale images (a) Original image with noise, (b) Cell
         detection by using Active Contour, (c) Better detection by using  the Novel
         algorithm.

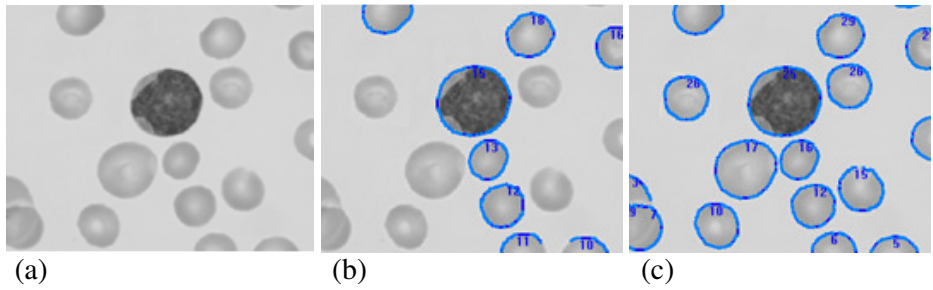

(a)　　　　　　(b)　　　　　　(c)

Figure 2. Example image from grayscale images (a) Original image, (b) Cell detection by
using Active Contour, (c) Better detection by using the Novel algorithm.

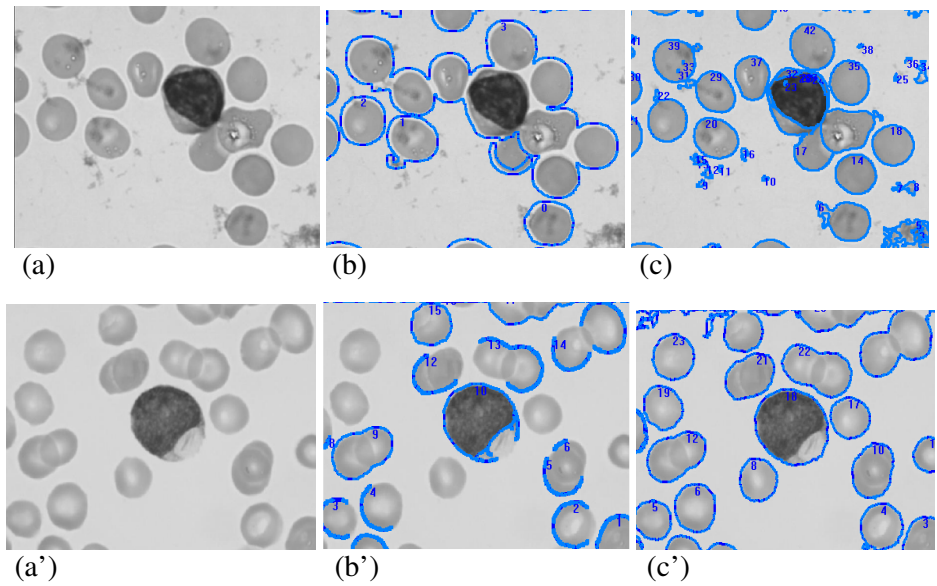

(a)　　　　　　(b)　　　　　　(c)

(a')　　　　　　(b')　　　　　　(c')

Figure 3. Example image from grayscale images (a, a') Original image, (b, b') Under
segmentation problem with Active contour, (c, c') Under segmentation solved by
using the Novel algorithm.

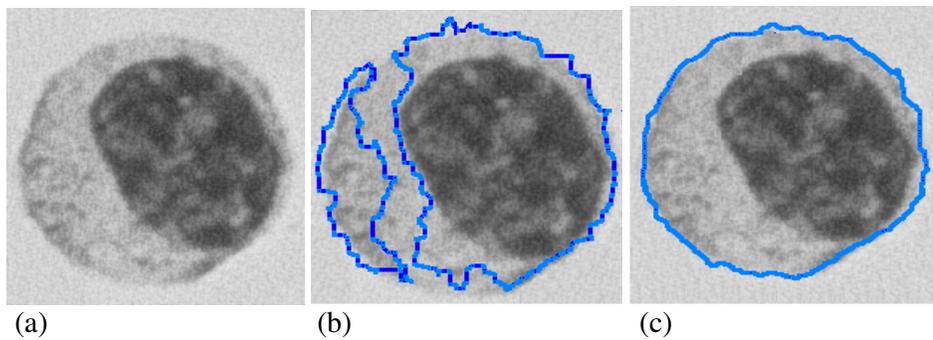

(a)　　　　　　(b)　　　　　　(c)

Figure 4. Example image from grayscale images (a) Over segmentation problem by using
Active contour  (b) Over segmentation problem solved, by using the Novel
algorithm.

# 5. CONCLUSION

Topological alignments and snakes are used in image processing, particularly in locating object boundaries. Both of them have their own advantages and limitations. Active contour, can locate the object boundaries dynamically and automatically from an initial contour. The main advantage of active contour models is the ability of the method to give a piece wise linear description of the shape of the object at the time of convergence, without extra processing. But the Active Contour models are strongly depends on finding strong image gradients to drive the contour. This significantly limits their utility, because frames with weak image boundaries and low contrast cause under and over segmentation which responsible for lacking accuracy. To solve this problem with the Active Contour model, we used Topological Alignments method to improve the performance of segmentation of cell tracking and increase the accuracy of cell detecting results.

In our experiments, we compared our algorithm with traditional snake. The results show that the algorithm can demonstrate the segmentation accuracy under weak image boundaries, under and over segmentation of leukocytes cells images, which the most cell detecting challenge problems and responsible for lacking accuracy in cell detecting techniques. The results show that the algorithm can deal with low contrast images and shape cells, demonstrate the segmentation accuracy under weak image boundaries, which responsible for lacking accuracy in image detecting techniques. We have achieved better segmentation and boundary detecting for the leukocytes cells images, also the ability of the system to improve the low contrast and deal with over and under segmentation.